%% file: acl_latex.tex
\pdfoutput=1

\documentclass[11pt]{article}

\usepackage[]{acl}

\usepackage{times}
\usepackage{latexsym}

\usepackage[T1]{fontenc}

\usepackage[utf8]{inputenc}

\usepackage{microtype}

\usepackage{inconsolata}
\usepackage{tikz}
\usepackage{comment}
\usepackage{booktabs}
\usepackage{amsmath,amssymb} 
\usepackage{color}
\usepackage[capitalize]{cleveref}
\usepackage{algpseudocode}
\usepackage{multirow}
\usepackage{multicol}
\usepackage{makecell}
\usepackage{enumitem}
\usepackage{float}
\usepackage{wrapfig}
\usepackage{bbding}
\usepackage[normalem]{ulem}
\usepackage{balance}
\usepackage{adjustbox}
\useunder{\uline}{\ul}{}
\usepackage{nameref}
\usepackage{graphicx}
\usepackage{subcaption}

\title{Less is More: Mitigating Multimodal Hallucination from \\an EOS Decision Perspective}

\author{Zihao Yue \\
  Renmin University of China \\
  \texttt{yzihao@ruc.edu.cn} \\\And
  Liang Zhang \\
  Renmin University of China \\
  \texttt{zhangliang00@ruc.edu.cn} \\\And
  Qin Jin\thanks{Corresponding Author.} \\
  Renmin University of China \\
  \texttt{qjin@ruc.edu.cn}}

\begin{document}
\maketitle

\begin{abstract}
\input{sections/0_abstract}
\end{abstract}
\input{sections/1_introduction}
\input{sections/2_eos-decision}
\input{sections/3_method}
\input{sections/4_related-works}
\input{sections/5_conclusion}

\section*{Acknowledgements}
We thank all reviewers for their insightful comments and suggestions. 
This work was partially supported by the Beijing Natural Science Foundation (No. L233008), the National Natural Science Foundation of China (No. 62072462), and the Outstanding Innovative Talents Cultivation Funded Programs 2023 of Renmin University of China.

\bibliography{custom}

\input{sections/6_appendix}

\end{document}

%% file: sections/0_abstract.tex
Large Vision-Language Models (LVLMs) often suffer from multimodal hallucinations, wherein they may create content that is not present in the visual inputs. In this paper, we explore a new angle of this issue: overly detailed training data hinders the model's ability to timely terminate generation, leading to continued outputs beyond visual perception limits. By investigating how the model decides to terminate generation with EOS, the special end-of-sentence token, we find that the model assesses the completeness of the entire sequence by comparing the generated text with the image. This observation suggests that the model possesses an inherent potential of making proper EOS decisions based on its visual perception to avoid overly lengthy outputs. To take advantage of such potential, we explore two methods to mitigate multimodal hallucinations: a training objective that enables the model to reduce hallucinations by learning from regular instruction data, and a data filtering strategy to prevent harmful training data from exacerbating model hallucinations. Both methods significantly improve the hallucination performance of LVLMs, without requiring any additional data or knowledge.\footnote{\url{https://github.com/yuezih/less-is-more}}


%% file: sections/1_introduction.tex
\input{figures/intro-1}

\section{Introduction}

Ever since Large Vision-Language Models (LVLMs)~\cite{lmm-survey} were achieved through bridging vision encoders with Large Language Models (LLMs)~\cite{gpt3,llm-survey}, they have been plagued by the problem of multimodal hallucinations, i.e., their text outputs may include unfaithful content to the visual inputs, such as non-existent objects~\cite{chair}, which greatly harms the reliability of their applications. Extensive research has shed light on the origins of multimodal hallucinations, including the inability of vision encoders to represent fine-grained visual details~\cite{comm,mmvp}, model reliance on inherent parametric knowledge such as language priors and statistical biases~\cite{vcd,lure}, and pervasive hallucinations in the training data itself~\cite{hallucidoctor,lrv-instruction}. In response to these insights, a variety of strategies have been proposed to mitigate hallucinations in LVLMs~\cite{woodpecker,halle-switch,ha-dpo}.

Although significant progress has been made, in this paper, we highlight a crucial but often overlooked source of hallucinations: the excessively detailed training data. For example, in the detailed image captioning task, the caption data for an image typically integrates rich visual semantics from multiple human annotations or vision expert models, and is rewritten into lengthy paragraphs by LLMs~\cite{llava}, as shown in \cref{fig:intro-1}. These training data, while high-quality and meeting our expectations for detail, may exceed the visual perception capability of LVLMs, especially for subtle image features such as small or easily confusable objects.
When trained with such data, in an attempt to fit the detail level and length distribution of ground truth captions, the model may risk expressing details that it cannot discern from the image, and therefore exhibit hallucinations.

Ideally, models should be trained to terminate generation upon reaching their visual perception limits to avoid hallucinations. However, because gauging such limits is not trivial, it is difficult to provide explicit supervision to teach models to stop generation timely, and it is impossible to construct training data that well matches model capabilities. Fortunately, we can draw inspiration from a closer examination of the model's decisions regarding the generation of the end-of-sentence (EOS) token. 

We first employ a saliency-based method to analyze how information flows from the context to the target position where the model predicts the next word. 
We discover that in predicting the EOS token, the model tends to rely more on all preceding sentences rather than just the current sentence. 
This leads to a hypothesis that \textbf{the model assesses the completeness of the entire sequence when deciding whether to terminate the generation}. 
Then, by manipulating the context, we observe that the model's tendency to predict EOS clearly varies depending on the semantic completeness of the generated text relative to the visual input. For instance, reducing visual context (easier to reach textual completeness) makes the model more likely to end the generation, whereas concealing textual context (further away from textual completeness) prompts continued generation. 
This confirms the hypothesis above and implies that such a {completeness assessment is accomplished by comparing the generated text with the perceived visual information}. 
These observations suggest that the model inherently holds the potential to make timely EOS decision to terminate the generation based on its visual perception. 
When the model decides to end the generation, it indicates that the current generated context sufficiently captures the visual information it can perceive, and any further outputs may exceed the model's visual perception limits, possibly leading to hallucinations.

To unlock such potential of models, we explore two approaches to enhance the model for better EOS decisions. (1) A learning objective for model training, termed \textbf{Selective EOS Supervision}. Simply modified from the Maximum Likelihood Estimation (MLE), it enables the model to mitigate hallucinations through learning from regular instruction data. It is applicable both for  further training to reduce hallucinations in existing models, and for initial instruction tuning~\cite{instructgpt} to alleviate the onset of hallucinations. Specifically, by briefly further training on the original instruction data, the sentence-level and instance-level hallucinations of LLaVA-1.5~\cite{llava1.5} are reduced by 26\% and 27\%, respectively. (2) A data filtering strategy based on \textbf{Scoring EOS Supervision}, to eliminate harmful training data that can impair the model's ability to end sequences. We design two metrics to assess the positive and negative impact of data on the model's EOS tendency, and combine them to rank and filter the training data. Experimental results show that removing a small portion of the data can significantly reduce the hallucinations of models trained on it. These findings further validate our hypothesis and provide simple yet effective solutions for mitigating multimodal hallucinations in LVLMs.

%% file: figures/intro-1.tex
\begin{figure}[t]
    \begin{center}
        \includegraphics[width=1\linewidth]{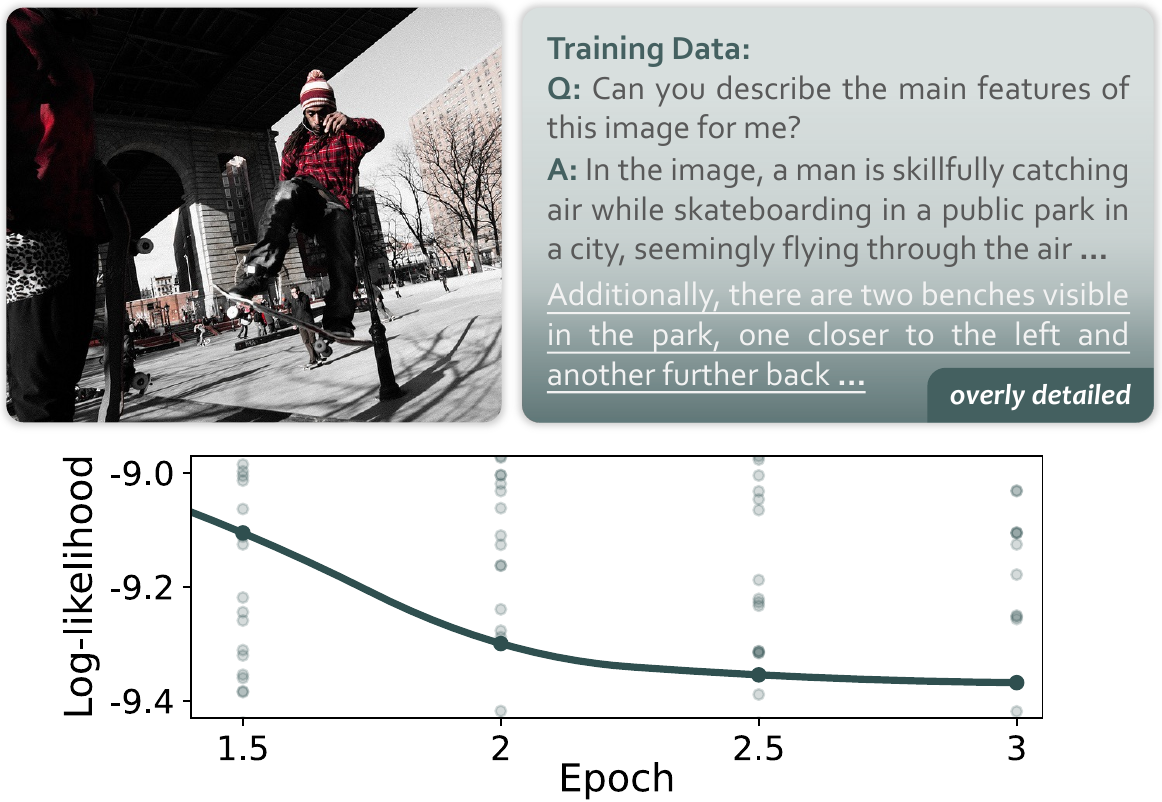}
    \end{center}
    \vspace{-8pt}
\caption{\textbf{Top:} An example from the LLaVA instruction data. The training data can be overly detailed to exceed the model's visual perception limits. \textbf{Bottom:} Average log-likelihood of the LLaVA {\scriptsize{(7b)}} model predicting EOS at positions labeled as EOS during instruction tuning. Training the model with overly detailed data leads to a decrease in its tendency to stop generation.}
\label{fig:intro-1}
\end{figure}

%% file: sections/2_eos-decision.tex
\section{EOS Decision}
\label{sec:analysis}

In autoregressive language models, sequences are completed through continuous next-token prediction (NTP)
. The termination of the process is achieved by introducing a special end-of-sentence (EOS) token, denoted as \(v_{EOS}\), into the vocabulary. At each NTP step, the model chooses between a regular content token and \(v_{EOS}\), deciding whether to continue the sequence generation or end it, which we refer to as EOS decision~\cite{eos}. 

In this section, we delve into how LVLMs reach EOS decisions. Specifically, in \cref{sec:analysis-1}, we analyze the contextual information that the model relies on to predict \(v_{EOS}\); in \cref{sec:analysis-2}, we explore how the model adjusts its tendency to terminate generation with the multimodal input. Corresponding findings are further discussed in \cref{sec:analysis-3}.

\subsection{Information Basis of EOS Decision}
\label{sec:analysis-1}

\input{figures/analysis-1}

We first investigate where the information for EOS decisions comes from in the context. Given that the context usually contains a long paragraph with multiple sentences, we group the context tokens into three parts: image tokens, preceding sentences, and the current sentence, to observe their respective contributions to the model's prediction decision. For comparison, we also examine the information for non-EOS target predictions occurring in the middle of a sequence, where the model needs to predict a regular content token. Since the context exposed for \(v_{EOS}\) prediction is the entire sequence, for a fair comparison, non-EOS targets are randomly selected from the last 10 tokens of the last sentence in the sequence. This ensures that they have access to all previous sentences while perceiving a sufficient portion of the current sentence.

We adopt the saliency score~\cite{saliency-1,saliency-2} as the metric for investigation. The saliency score of a token represents the sensitivity of the model to this token, i.e., how much its change affects the model prediction. As suggested by~\citet{labelwords}, we use the saliency score to quantify the information flow between tokens. Concretely, we feed the model with the first \(n\!-\!1\) tokens to predict the \(n\)-th target token through a forward pass, and obtain the cross-entropy loss \(\mathcal{L}(x)\) at the \(n\)-th target position. The saliency matrix \(I\) is given by 
\[I=\left|A\odot\frac{\partial{\mathcal{L}(x)}}{\partial{A}}\right|,\]
where \(A\) denotes the self-attention score matrix of the language model, \(\odot\) means element-wise product, and \(I(i,j)\) reflects the significance of the information flow from the \(j\)-th token to the \(i\)-th token. We compare the information flow patterns of EOS predictions and non-EOS predictions to elucidate the information basis of EOS decisions.

\noindent \textbf{Implementation details.} We choose the 7b version of LLaVA-1.5 as the model, containing a language decoder with 32 layers and 32 attention heads. The saliency matrix per layer is derived by averaging across all heads. The data used for investigation comes from Detail23K, a subset of the LLaVA-Instruction dataset~\cite{llava}, containing 23K detailed image descriptions for instruction tuning. For each run, we calculate the expectation over a random sample of 500 data entries.

\noindent \textbf{Results.} As illustrated in \cref{fig:analysis-1}, we first observe a pronounced information flow from contextual tokens to the target position, especially at higher layers (near the output). This implies a clear information aggregation pattern for model prediction. Then, we want to figure out where the information used for prediction comes from. As shown in \cref{fig:analysis-1} (left), for non-EOS predictions, the significance of information flows from the current sentence is comparable to that from previous sentences, despite the latter being significantly longer. This indicates that the current sentence is of great importance to model predictions. However, when the model is tasked with predicting \(v_{EOS}\), as depicted in \cref{fig:analysis-1} (right), the significance of information flows from previous sentences significantly increases and dominates. This suggests that the model, when predicting \(v_{EOS}\), places more emphasis on integrating information from all already generated content. This distinctive behavior indicates that the model's EOS decision is related to the current state of the entire sequence. Thus, we speculate that the model might be actively assessing the completeness of its text generation relative to its visual input, i.e., whether the current text is sufficient to describe its perceived visual information.

\subsection{Semantic Comparison for EOS Decision }
\label{sec:analysis-2}

To validate the hypothesis from \cref{sec:analysis-1}, we intervene in the multimodal input context and analyze the model's tendency for EOS predictions. Please note that the EOS decision does not solely occur at the final position of a sequence but at every position. However, for a well-trained language model, EOS predictions typically occur at the end of each sentence, i.e., the position right after the period. Hence, we focus on these target positions. We employ the same data and model mentioned in \cref{sec:analysis-1} for analysis, and obtain the conditional probabilities of \(v_{EOS}\) at various target positions through a forward pass. \cref{fig:analysis-2} (dotted line) illustrates the model's expected EOS tendency at each target position. A clear trend is that such a tendency increases as the sequence lengthens, implying the correlation between the textual richness and the EOS tendency. However, this correlation could stem from a variety of factors, for example, the length bias in training data can prompt the model to rely on positional embeddings for \(v_{EOS}\) predictions. To ablate potential disturbances, we additionally design three context manipulation methods:
\begin{itemize}[leftmargin=*, itemsep=0pt, parsep=0pt, before=\vspace{-0.5\baselineskip}, after=\vspace{-0.5\baselineskip}]
    \item \textbf{Visual reduction (\textit{image}\(-\)):} Applying a Gaussian noise mask to the input image, to reduce recognizable semantics in the image.
    \item \textbf{Visual augmentation (\textit{image}\(+\)):} Concatenating the image with a random one, to introduce visual information not described in the current text.\footnote{We also implement a variant that replaces the input image with a random new one instead of concatenation, to avoid increasing the absolute information richness (see \cref{sec:context-manipulation}).}
    \item \textbf{Textual reduction (\textit{text}\(-\)):} Using an attention mask to hide a portion of the exposed text. Here, we mask the first 30 tokens to ensure the coherence of the adjacent context for \(v_{EOS}\) predictions in the end part of the sequence.
\end{itemize}
These manipulation methods enable the augmentation or reduction of the multimodal contextual semantics without altering the sequence length. 

\input{figures/analysis-2}

\noindent \textbf{Results.} As illustrated in \cref{fig:analysis-2}, the reduction of image information through noise notably increases the model's tendency to predict \(v_{EOS}\). Conversely, introducing new image information or concealing text information, both implying a reduction in the relative textual completeness, lead to a decreased tendency of \(v_{EOS}\) prediction. These observations further support our conjecture that the model tends to assess the completeness of the current text to make an EOS decision, particularly, by comparing the generated text to the input image. Specifically, the more completely the image is described, the more likely the model is to terminate generation.

\subsection{Discussion}
\label{sec:analysis-3}

Our investigation on the information basis and the model's intrinsic criteria for EOS decisions reveal that models consider the current state of the entire sequence (\cref{sec:analysis-1}) and assess the completeness of generated text relative to the image (\cref{sec:analysis-2}). These findings suggest that while models may fit the training data length distribution and generate text beyond their capability limits, they still retain the inherent potential to adjust generation length according to visual perception. When the model tends to terminate generation, it can imply that the currently generated text adequately describes the visual information that the model can perceive. In \cref{sec:method}, we explore how this potential can be harnessed to mitigate multimodal hallucinations.

%% file: figures/analysis-1.tex


\begin{figure*}[ht]
    \centering
    \begin{subfigure}[b]{0.47\textwidth}
        \centering
        \includegraphics[width=\textwidth]{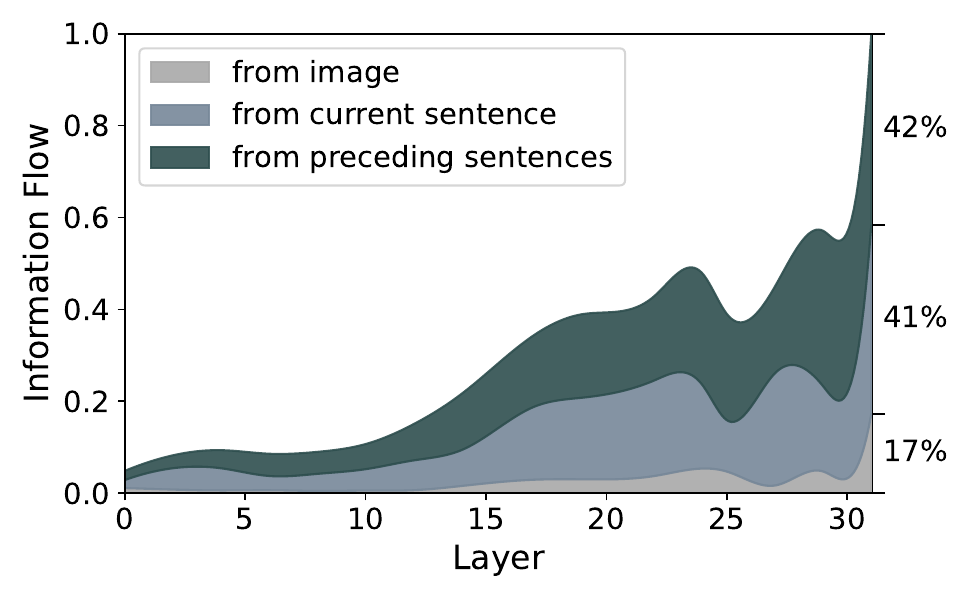}
        \vspace{-16pt}
        \label{fig:analysis-1a}
    \end{subfigure}
    \begin{subfigure}[b]{0.47\textwidth}
        \centering
        \includegraphics[width=\textwidth]{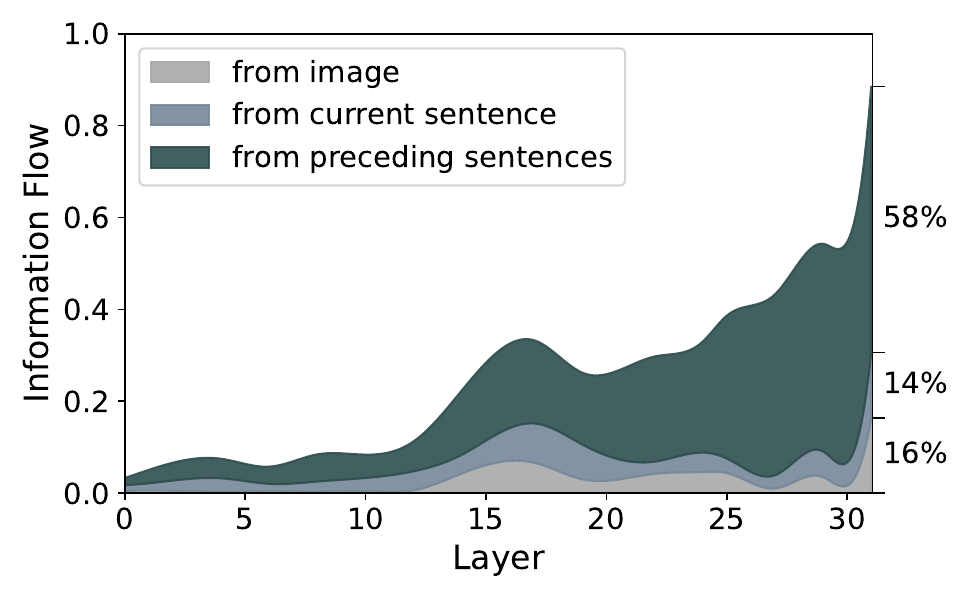}
        \vspace{-16pt}
        \label{fig:analysis-1b}
    \end{subfigure}
    \vspace{-4pt}
    \caption{Significance of the information flows from different parts of the input context to the target position during the prediction of a random token (left) and the EOS token (right). The significance refers to the proportion of these information flows out of a layer's total flows.}
    \label{fig:analysis-1}
\end{figure*}

%% file: figures/analysis-2.tex
\begin{figure}[t]
    \begin{center}
        \hspace*{-24pt}
        \includegraphics[width=0.9\linewidth]{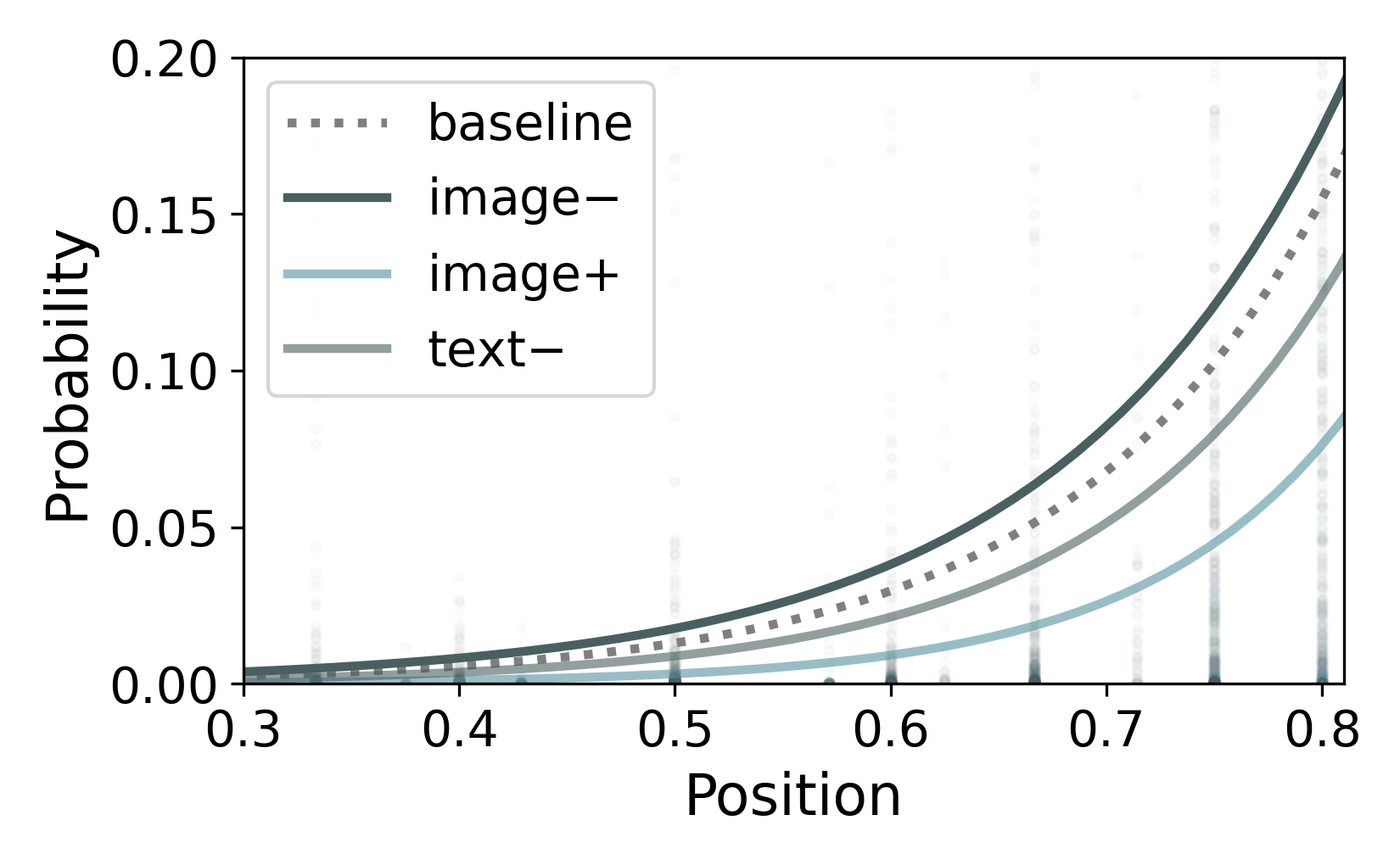}
    \end{center}
    \vspace{-12pt}
\caption{The predictive probability of \(v_{EOS}\) at various target positions, fitted by exponential functions. Position denotes the relative location \(i/N\) of the \(i\)-th target token among all \(N\) target tokens in the sequence.}
\label{fig:analysis-2}
\end{figure}

%% file: sections/3_method.tex
\section{Mitigating Multimodal Hallucinations}
\label{sec:method}

Inspired by the preceding analysis, we propose two approaches to mitigate multimodal hallucinations: (1) a learning objective, namely Selective EOS Supervision (\cref{sec:method-1}), which unlocks the model's capability to make EOS decisions at proper positions, thereby mitigating hallucinations; (2) a data filtering strategy, namely Scoring EOS Supervision (\cref{sec:method-2}), which eliminates training data that may hinder the model's capability to terminate generation in a timely manner.

\subsection{Selective EOS Supervision for Training}
\label{sec:method-1}

The instruction tuning of LVLMs typically utilizes Maximum Likelihood Estimation (MLE) as the training objective. Given the visual content \(v\) and previous tokens \(w_{<}\), the model predicts a probability distribution \(P^\mathcal{V}=\{p_1, p_2,\cdots,p_{|\mathcal{V}|}\}\) over the vocabulary \(\mathcal{V}\) to determine the next word, where \(p_j\) represents the probability of the \(j\)-th word in \(\mathcal{V}\). The model parameter \(\theta\) is optimized to maximize the likelihood of the label word indexed \(y\), with the loss function defined as:
\[\mathcal{L}_\text{MLE} = -\log(p_y|v,w_{<};\theta).\]
With such an objective, two optimization situations would happen regarding the \(v_{EOS}\) prediction: first, when the label is \(v_{EOS}\), the model's tendency to predict \(v_{EOS}\) will be enhanced; second, when the label is not \(v_{EOS}\), and if the model assigns some probability to \(v_{EOS}\), it will be penalized, becoming less likely to predict \(v_{EOS}\). 
Recalling our analysis in \cref{sec:analysis}, the model's tendency for  \(v_{EOS}\) prediction implies that the current text adequately represents its perceived visual information. Thus, in the second situation, stopping generation is the right choice. However, as the corresponding label is not \(v_{EOS}\) but a regular content token, the model will be discouraged from stopping and encouraged to continue generating content that may exceed its visual perception limits. Therefore, we aim to selectively preserve the first optimization situation, allowing the model to learn when to end generation, while minimizing the second optimization situation, to prevent compromising the model's EOS decision ability due to overly detailed training data.

\input{figures/method-1}
\input{tables/main-res}


To achieve the aforementioned goal, we implement a minor modification to MLE. Concretely, at positions where the label is not \(v_{EOS}\), we exclude \(v_{EOS}\) from the calculation of probability distribution. This means that the label's probability is determined using a modified softmax operation:
\[p_y = \mathrm{softmax}^*(\mathbf{z}_y) = \frac{\exp(\mathbf{z}_y)}{\sum_{j\in \mathcal{V}\setminus\{v_{EOS}\}}\exp(\mathbf{z}_j)},\]
where \(\mathbf{z}\) denotes logits. Since \(v_{EOS}\) does not participate in the probability distribution, it will not be suppressed by maximizing the label's probability, as depicted in \cref{fig:method-1}. This modification prevents MLE from undermining the model's inherent tendency to predict \(v_{EOS}\).
For positions where the label is \(v_{EOS}\), we retain vanilla MLE as objective, allowing the model to learn when to end sequences. 


\subsubsection{Experimental Settings}

\noindent \textbf{Models and datasets.} Our training objective can be applied to any LVLMs with an EOS token and optimized by MLE. As a representative, we select two widely used open-source LVLMs, LLaVA~\cite{llava} and MiniGPT~\cite{minigpt4} series. Among them, LLaVA, LLaVA-1.5~\cite{llava1.5}, and MiniGPT-v2~\cite{minigptv2} are trained with data recipes that include the LLaVA-Instruction dataset. Thus, we validate our method with these models by fine-tuning them on LLaVA-Instruction. Additionally, our experiment results show that a smaller subset of LLaVA-Instruction-150K which contains detailed captions, Detail23K, has a similar effect but brings significant computational efficiency. 
Thus, most of our experiments are conducted with Detail23K. If not specified, models undergo one epoch of training with LoRA~\cite{lora}. Other training details remain consistent with the models' official documentation.

\noindent \textbf{Evaluation.} Following previous works~\cite{opera,hallucidoctor}, we evaluate model hallucination with Caption Hallucination Assessment with Image Relevance (CHAIR)~\cite{chair}, a framework that quantifies object hallucination in image captions by comparing generated objects to the ground truth objects. The sentence-level score, CHAIR\(_S\), represents the proportion of captions that contain hallucinations, and the instance-level score, CHAIR\(_I\), denotes the frequency of hallucinated objects relative to all mentioned objects by the model. In addition, we measure an object recall to evaluate the semantic richness of generated captions. Our CHAIR tests involve 500 images randomly chosen from the MSCOCO validation set~\cite{mscoco}. 
We also adopt another metric FaithScore ~\cite{faithscore} to evaluate caption hallucination. It verifies the consistency between atomic facts in the caption and the input image with LLMs and visual expert models, for which we employ ChatGPT~\cite{chatgpt} and OFA~\cite{ofa}. It also provides a sentence-level score, FaithScore\(_S\). 

\noindent \textbf{Baselines.} Since our method facilitates models to timely terminate generation, which often results in shorter responses, we incorporate baselines that simply reduce the generation length, including sequence truncating and decoding with a length penalty. The truncating method keeps only the initial \(R\%\) of words in each caption, and the decoding method adopts an exponential length penalty to adjust the score of the EOS token during generation. Varying the truncating proportion or the length penalty factor leads to different generation length, and affects both hallucination and recall performance.
Additionally, we include two recently proposed plug-in methods: (1) Visual Contrastive Decoding (VCD)~\cite{vcd}, which contrasts the output distributions derived from the original and noisy visual inputs, to reduce the influence of the model's parametric knowledge. (2) Over-Trust Penalty and Retrospection-Allocation (OPERA)~\cite{opera}, a decoding strategy that penalizes the model's over-reliance on certain tokens and allows roll-back when needed. We test VCD at different noise steps of 200, 500, 700, and 999, and report the optimal results. For OPERA, our implementation follows the suggestions in their released code, including two hyperparameter configurations for standard and fast inference.

\input{figures/chair-recall}
\subsubsection{Results}

\noindent \textbf{Versus the original model.} As shown in \cref{tab:main-res}, after a single training epoch on the detailed caption subset, 
Detail23K, using our learning objective, all models tend to produce shorter captions and notably reduce hallucinations at both the sentence and instance levels (\(r_1\) vs \(r_6\), \(r_{7}\) vs \(r_{8}\), etc.).
This improvement is even more significant when using the full 150K instruction data (\(r_1\) vs \(r_5\)), resulting in a 26.4\% and 26.6\% decrease in CHAIR\(_S\) and CHAIR\(_I\) of LLaVA-1.5 {\footnotesize{(7b)}}, respectively. While our method leads to some decrease in recall (e.g., \(-\)1.8\% of LLaVA-1.5 {\footnotesize{(7b)}} and \(-\)2.6\% of LLaVA-1.5 {\footnotesize{(13b)}}), we view this as a beneficial compromise since the models become more conservative and less likely to ``guess'' uncertain visual content. More analysis can be found in \cref{sec:appendix-method-1}.

\input{figures/experiment-1}

\noindent \textbf{Versus baselines.} As demonstrated in \cref{fig:chair-recall}, the truncating and length-penalty decoding baselines, with varying length-controlling configurations, effectively reduce hallucinations at the cost of Recall. However, these methods fall short of ours for either being less effective in alleviating hallucinations or resulting in more significant recall loss. Our method also outperforms existing methods, i.e., VCD and OPERA, as shown in \cref{tab:main-res}. Our proposed method does not require additional data construction or other expert models. Furthermore, unlike decoding methods, it does not slow down inference and remains technically compatible with various decoding strategies. Therefore, it presents a viable, practical supplement or alternative to current methods.

\noindent \textbf{Versus MLE.} To confirm that the improvement in model hallucination performance results from our modification to MLE, we also conduct a comparison by further training the model using the vanilla MLE. As shown in \cref{fig:experiment-1}, the performance of the model optimized by MLE varies throughout the training and remains at the original level. This variation suggests that different training samples impact the model differently; some may enhance the model's EOS tendency while others do the opposite, collectively preserving the model's initial generation habits. In contrast, with our modified learning objective, the degree of model hallucination steadily decreases throughout one epoch of training, with the model eventually significantly outperforming its MLE counterpart. This indicates that our selective supervision consistently enhances the model's EOS decision with varying data inputs.

\noindent \textbf{Instruction tuning from scratch.} Beyond further training existing models, our method is also compatible with instruction-tuning new models, starting from a vision-language aligned yet not instruction-tuned state. \cref{tab:from-scratch} presents the results of fine-tuning LLaVA for 3 epochs with LLaVA-Instruction-150K. With our learning objective, the model's sentence-level and instance-level hallucinations are reduced by 31.6\% and 15.9\%, respectively. Combining our learning objective with the vanilla MLE at a 1:1 ratio achieves a more balanced performance between hallucinations and recall.

\input{tables/from-scratch}

\subsection{Scoring EOS Supervision for Training Data Filtering}
\label{sec:method-2}


As the preceding analysis shows, learning from overly detailed data can impair a model's ability to predict \(v_{EOS}\), so an intuitive solution is to filter out such ``harmful'' training data. 

As described in \cref{sec:method-1}, there exist two optimization situations regarding \(v_{EOS}\) prediction, corresponding to a positive effect that enhances the model's EOS tendency when the label is \(v_{EOS}\) and a negative effect otherwise. We thus design two metrics to quantitatively evaluate the two effects on models when trained with a certain data sample:
\[
\scalebox{0.97}{
$\begin{aligned}
    S_{pos}\!&\!= -\!\sum_{i=1}^{N} [y_i\!=\!v_{EOS}] \log(p_{v_{EOS}}|v,w_<;\theta^*); \\
    S_{neg}\!&\!= -\!\sum_{i=1}^{N} [y_i\!\neq\!v_{EOS}] \log(1\!-\!p_{v_{EOS}}|v,w_<;\theta^*).
\end{aligned}$
}
\]
Here, \(\theta^*\) is a reference model used for evaluating the data. For positions where the label is \(v_{EOS}\), we define \(S_{pos}\) as the cross-entropy loss of the reference model predicting the label. A large cross-entropy loss indicates that, on this particular training data, the model fails to predict \(v_{EOS}\) to end the sequence, and the feedback from the training loss will enhance the model to learn this capability. Thus, \(S_{pos}\) quantifies the positive effect of the data on the model's EOS prediction. Conversely, for positions where the label is not \(v_{EOS}\), if the model tends to predict \(v_{EOS}\), this tendency will be undesirably suppressed. Particularly, a larger \(p_{v_{EOS}}\) leads to a more significant negative effect, especially when \(p_{v_{EOS}}\) approaches 1. Therefore, as defined above, \(S_{neg}\) serves to estimate the negative effect of the data on the model's EOS decision.

Intuitively, our goal is for \(S_{pos}\) to be as high as possible, indicating strong penalties for the model's inability to predict \(v_{EOS}\) where it should, and for \(S_{neg}\) to be as low as possible, reflecting minimal suppression on the model's EOS tendency. Therefore, we calculate a composite score \(S_{final}\!=\!S_{neg}\!-\!S_{pos}\) to estimate the ``harmfulness'' of the data. It is recommended to remove the highest-scoring data parts from training to achieve a more desired outcome for appropriate EOS decisions. With the shared goal of preserving the model's EOS decision capability, our data filtering strategy can serve as an alternative to the Selective EOS Supervision described in \cref{sec:method-1}.

\subsubsection{Experimental Settings}

\noindent \textbf{Data filtering.} We apply the proposed data filtering strategy to the LLaVA-Instruction-150K dataset. The model used for scoring \(S_{final}\) is the instruction-tuned version of LLaVA-1.5 {\scriptsize{(7b)}}. 
We test three data filtering ratios, ranging from 10\% to 30\%, to remove data with the highest \(S_{final}\). Additionally, we evaluate a random filtering strategy where 20\% of the data is removed randomly, as well as a reversed filtering strategy, with 20\% of data with the lowest \(S_{final}\) being removed. 

\noindent \textbf{Models.} We fine-tune the LLaVA {\scriptsize{(7b)}} model from scratch with the filtered data to validate their effectiveness. Following common practice, all models are trained for 3 epochs with a batch size of 128. We adopt QLoRA~\cite{qlora} to reduce computational load. Note that in this subsection, models are trained with the vanilla MLE. 

\input{tables/filter}

\subsubsection{Results}

As shown in \cref{tab:filter}, by removing a small proportion, i.e., 20\%, of ``harmful'' data from the original training set, the model significantly reduces learning hallucinations during instruction tuning, resulting in the sentence-level and instance-level hallucinations reduced by 23.7\% and 23.2\%, respectively.
In contrast, the reversed filtering which removes the least ``harmful'' data leads to opposite effects, greatly exacerbating the model hallucination, while the random removal brings no improvements in sentence-level hallucination performance.  This shows that our criteria used for data filtering well reflect the impact of the data on the model's ability to end generation.

Another interesting finding is that filtering the data does not bring a big change in the length of the training data, but it does significantly affect the length of the model generation. For instance, the reversed filtering leaves the average length of the training data almost unchanged, but the average length of model generation nearly doubles. This implies that the impact of our data filtering strategy does not come from changing the length distribution of the training data. Instead, it affects the model through manipulating the EOS supervision, further validating our motivation.

\subsection{Discussion}

In this section, to fully exploit the potential of the model to properly end generation according to its visual perception, we propose to mitigate hallucinations by not suppressing its inherent EOS tendency, through a training objective (\cref{sec:method-1}) and a data filtering strategy (\cref{sec:method-2}). From a practical perspective, the former approach has the merits of broader applicability and easier deployment, especially when used to further train existing models. The latter has greater compatibility since filtered data can be paired with various training methods.

%% file: figures/method-1.tex
\begin{figure}[t]
    \begin{center}
        \includegraphics[width=1\linewidth]{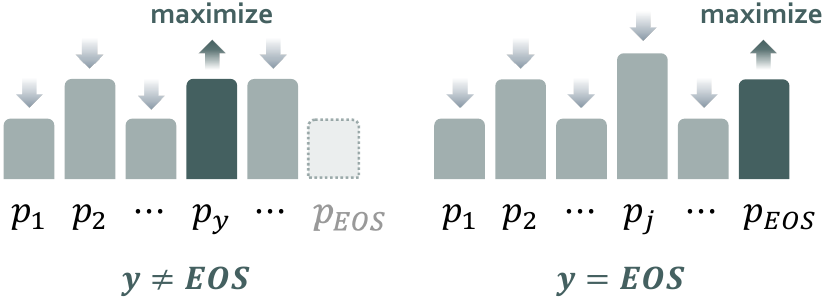}
    \end{center}
    \vspace{-4pt}
\caption{Illustration of the probability distribution derived from our proposed Selective EOS Supervision. Arrows indicate the maximizing and minimizing effects of the training objective on the probability of each word. When the label is not EOS, the EOS token is excluded from the probability distribution.}
\label{fig:method-1}
\end{figure}

%% file: tables/main-res.tex
\begin{table*}[t]
\caption{\label{tab:main-res}
Hallucination performance of different models. \textbf{w/ Cap.} and \textbf{w/ Inst.} denote fine-tuning models with the detailed caption subset, Detail23K, and with the full LLaVA-Instruction-150K, respectively. \textbf{Faith}: FaithScore.
}
\vspace{-4pt}
\centering
\small
\begin{tabular}{@{}c|l|l|c|ccc|cc@{}}
\toprule
Row & Model & Method & Length & CHAIR\(_S\) $\downarrow$ & CHAIR\(_I\) $\downarrow$ & Recall $\uparrow$ & Faith $\uparrow$ & Faith\(_S\) $\uparrow$ \\ \midrule
1 & \multirow{6}{*}{LLaVA-1.5 \scriptsize{(7b)}} & - & 100.6 & 50.0 & 15.4 & {\ul 77.1} & 87.0 & 68.8 \\
2 &  & VCD & 100.4 & 48.6 & 14.9 & \textbf{77.3} & 87.1 & 70.2 \\
3 &  & OPERA & 98.6 & 47.8 & 14.6 & 76.8 & 88.0 & {\ul 72.6} \\
4 &  & OPERA \scriptsize{(fast)} & 85.3 & 48.6 & 14.5 & 76.7 & 87.7 & 71.3 \\
5 &  & \textbf{Ours \scriptsize{(w/ Inst.)}} & 76.2 & \textbf{36.8} & \textbf{11.3} & 74.3 & {\ul 88.4} & \textbf{73.0} \\
6 &  & \textbf{Ours \scriptsize{(w/ Cap.)}} & 79.7 & {\ul 40.2} & {\ul 12.3} & 75.7 & \textbf{89.3} & 72.3 \\ \midrule
7 & \multirow{2}{*}{LLaVA-1.5 \scriptsize{(13b)}} & - & 100.9 & 47.2 & 13.0 & \textbf{77.3} & 87.6 & \textbf{73.1} \\
8 &  & \textbf{Ours \scriptsize{(w/ Cap.)}} & 85.1 & \textbf{36.8} & \textbf{11.4} & 75.3 & \textbf{88.8} & 72.8 \\ \midrule
9 & \multirow{2}{*}{LLaVA \scriptsize{(7b)}} & - & 57.8 & 35.4 & 13.8 & \textbf{64.8} & 86.9 & 67.4 \\
10 &  & \textbf{Ours \scriptsize{(w/ Cap.)}} & 39.9 & \textbf{27.0} & \textbf{13.2} & 57.1 & \textbf{88.9} & \textbf{71.6} \\ \midrule
11 & \multirow{2}{*}{MiniGPTv2 \scriptsize{(7b)}} & - & 87.2 & 38.0 & 11.1 & 66.3 & 85.6 & 67.8 \\
12 &  & \textbf{Ours \scriptsize{(w/ Cap.)}} & 62.2 & \textbf{27.0} & \textbf{9.8} & \textbf{66.6} & \textbf{89.9} & \textbf{76.0} \\ \bottomrule
\end{tabular}
\end{table*}

%% file: figures/chair-recall.tex
\begin{figure}[t]
    \begin{center}
        \hspace*{-12pt}
        \includegraphics[width=0.83\linewidth]{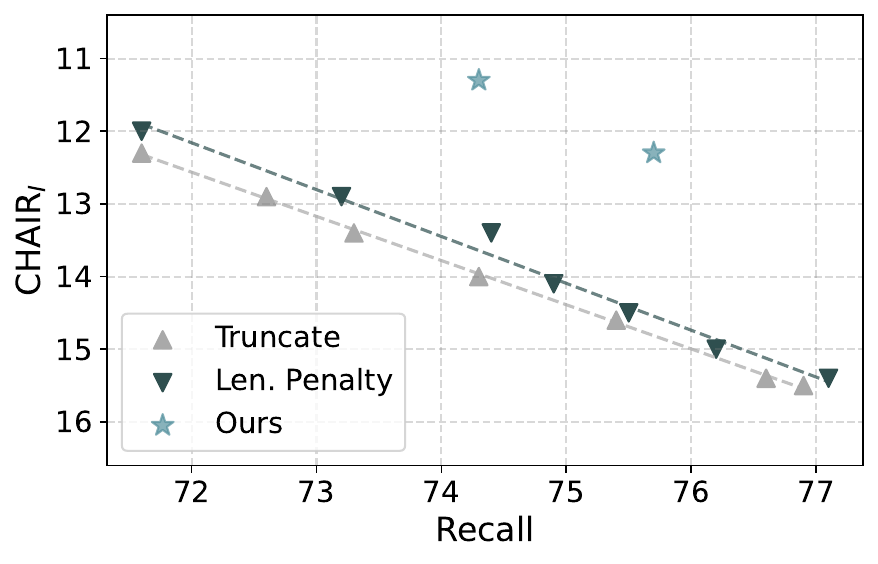}
    \end{center}
    \vspace{-6pt}
\caption{Hallucination vs. Recall performance of LLaVA-1.5 {\scriptsize{(7b)}}. \textbf{Ours}: the models fine-tuned on \textbf{Inst.} and \textbf{Cap.} respectively with our training objective.}
\label{fig:chair-recall}
\end{figure}

%% file: figures/experiment-1.tex
\begin{figure}[t]
    \begin{center}
        \includegraphics[width=0.9\linewidth]{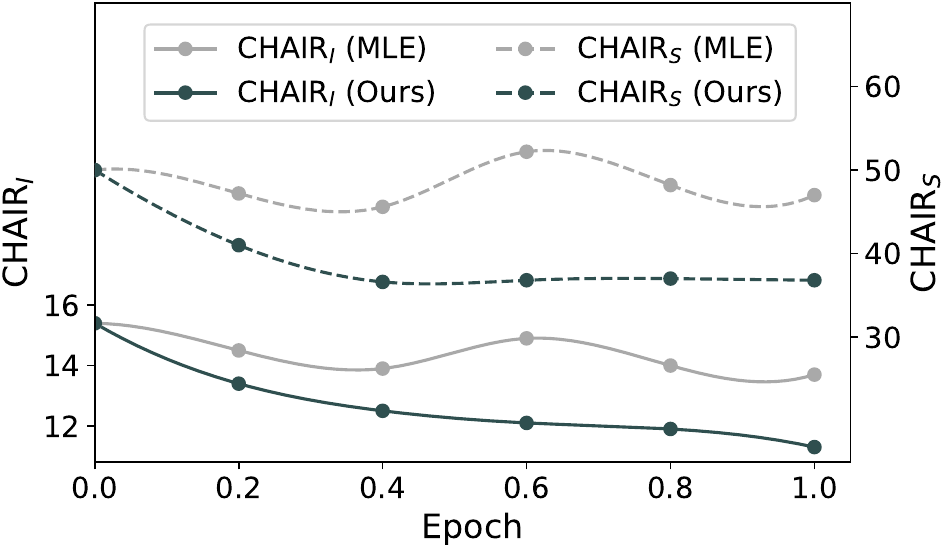}
    \end{center}
    \vspace{-6pt}
\caption{CHAIR performance trends of LLaVA-1.5 {\scriptsize{(7b)}} throughout training on LLaVA-Instruction-150K.}
\label{fig:experiment-1}
\end{figure}

%% file: tables/from-scratch.tex
\begin{table}[t]
\caption{\label{tab:from-scratch}
CHAIR evaluation results of the LLaVA {\scriptsize{(7b)}} models instruction-tuned from scratch.
}
\vspace{-4pt}
\centering
\small
\begin{tabular}{@{}l|cccc@{}}
\toprule
Loss & Length & CHAIR\(_S\) & CHAIR\(_I\) & Recall \\ \midrule
MLE & 57.8 & 35.4 & 13.8 & \textbf{64.8} \\
Ours & 36.1 & \textbf{24.2} & \uline{11.6} & 55.9 \\
Combined & 42.7 & \uline{26.6} & \textbf{11.0} & \uline{57.5} \\ \bottomrule
\end{tabular}
\end{table}

%% file: tables/filter.tex
\begin{table}[t]
\caption{\label{tab:filter}
CHAIR evaluation results of models trained with different data. C\(_{S\mathrm{/}I}\) denotes CHAIR\(_{S\mathrm{/}I}\).
}
\vspace{-4pt}
\centering
\small
\begin{tabular}{@{}c|cc|cccc@{}}
\toprule
\multirow{2}{*}{Row} & \multicolumn{2}{c|}{Train. Data} & \multicolumn{4}{c}{Model Performance} \\ \cmidrule(l){2-7} 
 & Filter & Len. & Len. & C\(_S\) & C\(_I\) & Recall \\ \midrule
1 & \multicolumn{1}{l}{Original} & 178.3 & 57.8 & 35.4 & 13.8 & {\ul 64.8} \\ \midrule
2 & 10\% & 171.7 & 63.7 & 35.4 & 14.0 & 64.5 \\
3 & 20\% & 168.2 & 45.5 & \textbf{27.0} & \textbf{10.6} & 58.9 \\
4 & 30\% & 166.7 & 49.2 & {\ul 29.4} & {\ul 11.7} & 58.0 \\ \midrule
5 & \multicolumn{1}{l}{Random} & 178.2 & 68.9 & 35.5 & 11.8 & 61.9 \\
6 & \multicolumn{1}{l}{Reversed} & 176.8 & 100.6 & 46.6 & 18.9 & \textbf{68.6} \\ \bottomrule
\end{tabular}
\end{table}

%% file: sections/4_related-works.tex
\section{Related Works}

\noindent \textbf{Hallucination Origins.} Investigations on the causes of multimodal hallucination in LVLMs identify three main factors: (1) Limited visual representations. For example, the visual encoders commonly employed in LVLMs depict abstract features while struggling to capture fine-grained visual details~\cite{comm, mmvp, haelm}. \citet{hacl} observe a modality gap persists between visual and textual features, despite vision-language alignment. (2) Models' over-reliance on parametric knowledge, such as statistical biases and language priors, rather than on visual evidence~\cite{lure, vcd, lrv-instruction, halle-switch, hallusionbench}. (3) Inferior data for instruction tuning. This includes insufficient visual supervision~\cite{visual-supervision}, a lack of positive/negative human feedback~\cite{rlhf-v}, and the presence of hallucinations within the training data~\cite{hallucidoctor, lrv-instruction}. Our paper identifies a new source of hallucinations: overly detailed training data hinders the model's inherent EOS decision ability, further enriching existing explanations.

\noindent \textbf{Mitigation Solutions.} An effective way to reduce hallucinations is to construct high-quality \textbf{data}, including employing automatic data cleaning pipelines~\cite{hallucidoctor}, generating~\cite{lrv-instruction} or rewriting~\cite{recaption} training data with LLMs, and integrating human feedback into annotations~\cite{mhaldetect}. \textbf{Training} approaches view hallucinatory data as negative examples, and adopt preference optimization~\cite{halle-switch, ha-dpo, llava-rlhf, rlhf-v, silkie} or contrastive learning~\cite{hacl} to enhance models' resistance to hallucinations. \textbf{Inference} strategies focus on the decoding process, suppressing models' reliance on parametric biases~\cite{vcd} or penalizing inferior attention patterns~\cite{opera}. Other works explore \textbf{posthoc-fixing} ways to rectify hallucinations in model outputs, by training a revisor model~\cite{lure}, employing expert models~\cite{woodpecker}, and prompting the original model for self-correction~\cite{volcano}. In this paper, we propose a new learning objective and a data filtering strategy, belonging to the training and data perspectives, respectively.

%% file: sections/5_conclusion.tex
\section{Conclusion}

This paper investigates the multimodal hallucination issue in large multimodal models. We suggest that overly detailed training data can prevent the model from stopping generation at the appropriate time, thus leading to hallucinated outputs. By examining the model's inner behavior of EOS prediction, we discover that the model inherently holds the potential to terminate generation based on its visual perception limits. To enhance such potential, we develop two approaches, a learning objective for training models and a data filtering strategy for selecting training data, both of which facilitate the model learning to timely terminate generation and significantly reduce hallucinations.

\section*{Limitations}
\label{sec:limitation}

This work presents a novel perspective on the origins of multimodal hallucinations in large multimodal models with corresponding solutions. However, it faces several limitations. First, it focuses solely on generative tasks, i.e., detailed image description, without covering hallucinations in broader tasks like classification-oriented Visual Question Answering (VQA). 
Second, our solutions are examined only on multimodal models, though technically they could also be applied to unimodal large language models. We leave this possibility for future exploration. Third, our solutions mitigate hallucinations by enhancing the model's ability to timely conclude sequences. While effective, they address only the simplest source among various causes of hallucinations. Fully solving the problem of hallucination remains a substantial challenge.

\section*{Ethics Statement}

This work focuses on reducing hallucinations in large multimodal models to enhance their reliability and trustworthiness. We have carefully considered the ethical implications of our work and anticipate no significant ethical concerns. This work was carried out using publicly available and commonly used data and models, and our findings may inherit the biases and limitations carried in these resources.

%% file: sections/6_appendix.tex

\appendix
\input{tables/appendix-computation}

\section{Additional Implementation Details}

\subsection{Experiment in Figure 1}

For the trend depicted in \cref{fig:intro-1}, the model undergoes fine-tuning on the LLaVA-Instruction-150K dataset over 3 epochs and is evaluated with the same data used in \cref{sec:analysis}. At the initial stages of training, the model shows notable fluctuations in performance due to the lack of prior fitting to the instruction data. To better demonstrate how data affects the model's EOS tendency, we focus on the latter half of the training period, where performance begins to stabilize.

\subsection{Context Manipulation}

In \cref{sec:analysis-2}, we reduce the semantics of images by overlaying them with a Gaussian noise mask. This involves gradually introducing minor amounts of Gaussian noise over \(T\) steps, mirroring the forward diffusion process used in image generation tasks~\cite{diffusion}. Our implementation follows that of~\citet{vcd}, and we set \(T\) to 500 for analysis.

\subsection{Computation}

\cref{tab:computation} presents the computational cost of model training on a setup with 8 NVIDIA RTX A6000 GPUs. Our proposed training objective and data filtering strategy do not introduce a noticeable increase in training costs. In this work, all experimental results are derived from single runs, with greedy search as the decoding strategy.

\section{Additional Results}

\subsection{Information Aggregation Pattern}

\input{figures/appendix-information-flow}

In \cref{sec:analysis-1}, we observe a significant information aggregation from the context to the target position during token prediction. This section further clarifies the details of this information aggregation pattern. At lower layers (near the input), the information from regular content tokens within a sentence converges at the sentence's end, typically a period, seemingly summarizing the entire sentence. At higher layers (near the output), this ``summarized'' information then aggregates to the target position for the next token prediction. Following~\citet{labelwords}, we illustrate these effects in \cref{fig:appendix-information-flow}, where such effects occur for both EOS and non-EOS predictions. This observation closely aligns with the findings by~\citet{labelwords} in in-context learning (ICL), where the labels of in-context demonstrations act as "anchors" that aggregate information at lower layers and provide it for the final prediction at higher layers. This hierarchical information aggregation pattern elucidates how information moves within contexts and underpins our analysis in \cref{sec:analysis-1}. We hope these observations 
can shed some light on future research.


\input{figures/appendix-context-manipulation}
\subsection{Context Manipulation}
\label{sec:context-manipulation}

In \cref{sec:analysis-2}, we design three context manipulation methods to analyze how the model adjusts its EOS tendency according to these interventions. In addition to these methods, we also implement a variant of visual augmentation (\textit{image}\(+\)), where we replace the input image with a random new one instead of concatenating a random image with the input image. This method can also decrease the relative completeness of the text, while not necessarily increasing the absolute information richness. The results in \cref{fig:appendix-context-manipulation} demonstrate a similar impact from both variants, suggesting that the model does not merely compare the absolute semantic richness of the text and the image, but assesses the relative semantic completeness of the text to the image, i.e., whether the existing text encompasses the perceived visual information. This observation further supports our conjecture.

\input{figures/appendix-eos-tendency}
\input{tables/appendix-omitted}
\input{figures/appendix-score-distribution}

\subsection{Selective EOS Supervision}
\label{sec:appendix-method-1}

\noindent \textbf{EOS prediction tendency.} In \cref{fig:appendix-eos-tendency}, we illustrate the EOS prediction tendency (average probability) of the LLaVA-1.5 {\scriptsize{(7b)}} model during further training on Detail23K. With Selective EOS Supervision proposed in \cref{sec:method-1}, the model's tendency to predict EOS rises and stabilizes, while the model optimized by MLE shows no change in this behavior. This suggests that the proposed training objective effectively helps the model regain its capability to timely conclude sequences.

\noindent \textbf{Dissecting omitted content.} As our method reduces the generated content to alleviate hallucinations, it is interesting to investigate what is ``omitted'' by our method from the originally generated captions, specifically, how many ``omitted'' objects are correct and how many are hallucinations. We extract the generated objects from the outputs of both the original model and our further trained models, using the same technique as in the CHAIR evaluation. Then, we focus on these objects that are mentioned by the original model but not by our models, which are ``omitted'' from the original captions. As the results of the \textit{Halluc. Rate} (hallucinated object rate of omission) in \cref{tab:omitted} shows, nearly 3/4 of the ``omitted'' objects are hallucinations, implying that such an omission is beneficial. 

Furthermore, we analyze the average counts of correct and hallucinated objects in the model generation, as a supplement to the CHAIR metrics, to demonstrate more comprehensively how our method impacts the quality of model generation. As shown in \cref{tab:omitted-2}, our method reduces hallucinations while largely preserving the correct content.

\input{tables/appendix-mme-pope}

\subsection{Scoring EOS Supervision}
\label{sec:appendix-method-2}

\noindent \textbf{Data Score Distributions.} In \cref{sec:method-2}, we discuss two metrics, \(S_{pos}\) and \(S_{neg}\), which are summed over positions labeled as EOS and non-EOS, respectively. A natural concern is that the number of non-EOS positions far exceeds that of EOS positions, raising the question of whether the combined \(S_{final}\) might be dominated by \(S_{neg}\). To clarify this, we examine the score distributions within the LLaVA-Instruction-150K dataset. As illustrated in \cref{fig:appendix-score-distribution}, the value distributions of \(S_{pos}\) and \(S_{neg}\) are comparable in magnitude, and the \(S_{final}\) distribution is approximately normal with zero mean. This indicates a balance between \(S_{pos}\) and \(S_{neg}\) with neither metric dominating, and the top \(S_{final}\) scores necessitate both high \(S_{neg}\) and low \(S_{pos}\). Thus, by maintaining a straightforward formulation of \(S_{final}\!=\!S_{neg}\!-\!S_{pos}\) without introducing a balancing hyperparameter, the contributions of both metrics are reflected. Initial experiments also reveal that relying solely on \(S_{neg}\) for data filtering increases hallucinations, as it can lead to mistakenly removing data with high \(S_{pos}\); whereas using \(S_{pos}\) alone does reduce hallucinations, but is not as effective as \(S_{final}\) and will bring greater recall loss. Balancing both metrics yields the most desirable outcomes.

\subsection{MME and POPE Evaluation}

As mentioned in \nameref{sec:limitation}, our proposed techniques focus on mitigating hallucinations in generative tasks by adjusting the models' propensity for appropriately concluding outputs. However, these methods are not directly transferable to addressing hallucination problems in broader Visual Question Answering (VQA) tasks, such as those evaluated in the MME~\cite{mme} and POPE~\cite{pope} benchmarks. The MME benchmark assesses the model's capabilities in terms of perception and cognition, whereas POPE concentrates on object hallucinations. Both benchmarks challenge models with \textit{Yes-or-No} questions. As shown in \cref{tab:appendix-mme-pope}, our methods do not yield performance gains on these benchmarks. The effectiveness of our approaches in generative tasks suggests that a model's failure to timely stop generation is an important hallucination source. However, addressing this issue alone does not fundamentally solve all hallucination problems as the origins of multimodal hallucinations are multifaceted. This area remains open for further investigation.

\subsection{Qualitative Results}
\balance

\input{figures/appendix-case}

We present qualitative examples of our methods, Selective EOS Supervision in \cref{fig:appendix-case-1} and Scoring EOS Supervision in \cref{fig:appendix-case-2}. The baseline models often produce hallucinations towards the end of their outputs, as they try to include too many details from the image, sometimes beyond their visual perception limits. This also explains why simply truncating sequences can reduce hallucinations. However, with our methods, the models better retain the innate ability to stop generation right after covering what they can visually perceive. This prevents the generation of overly lengthy, inaccurate, or irrelevant outputs that lower the overall quality and information density of the generated content, echoing the principle that ``less is more.''

%% file: tables/appendix-computation.tex
\begin{table*}[t]
\caption{\label{tab:computation}
The computational burden of model training. \textbf{Train. Strategy} means either further training models that have already been fine-tuned for instruction following or instruction tuning from scratch. \textbf{PEFT} stands for the Parameter-Efficient Fine-Tuning (PEFT) strategies we adopt.
}
\vspace{-4pt}
\centering
\small
\begin{tabular}{@{}cclccc@{}}
\toprule
Train. Strategy & PEFT & Model & Data & Epoch(s) & Total GPU Time \\ \midrule
\multirow{5}{*}{Further} & \multirow{5}{*}{LoRA} & LLaVA-1.5 \scriptsize{(7b)} & Cap. & 1 & $\sim$0.6 h \\
 &  & LLaVA-1.5 \scriptsize{(7b)} & Inst. & 1 & $\sim$9.0 h \\
 &  & LLaVA-1.5 \scriptsize{(13b)} & Cap. & 1 & $\sim$3.0 h \\
 &  & LLaVA \scriptsize{(7b)} & Cap. & 1 & $\sim$0.6 h \\
 &  & MiniGPTv2 \scriptsize{(7b)} & Cap. & 1 & $\sim$2.5 h \\ \midrule
From-scratch & QLoRA & LLaVA \scriptsize{(7b)} & Inst. & 3 & $\sim$11.0 h \\ \bottomrule
\end{tabular}
\end{table*}

%% file: figures/appendix-information-flow.tex
\begin{figure}[t]
    \begin{center}
        \includegraphics[width=0.9\linewidth]{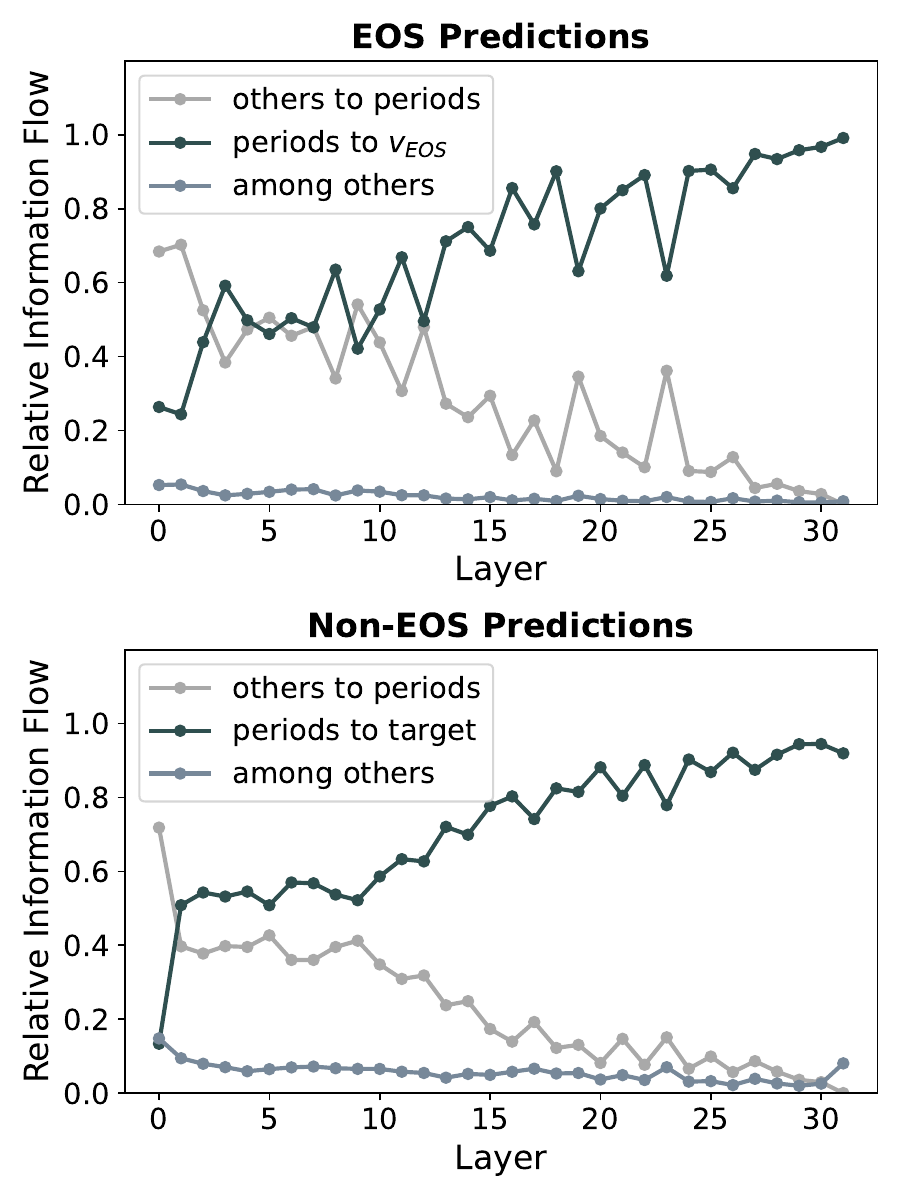}
    \end{center}
    \vspace{-8pt}
\caption{Relative significance of the information flow from regular content tokens (others) to periods, from periods to the target position for prediction, and among others. The significance is averaged over information flow targets and normalized across these three aspects for clearer comparison.}
\label{fig:appendix-information-flow}
\end{figure}

%% file: figures/appendix-context-manipulation.tex
\begin{figure}[t]
    \begin{center}
        \vspace{-1pt}
        \includegraphics[width=0.9\linewidth]{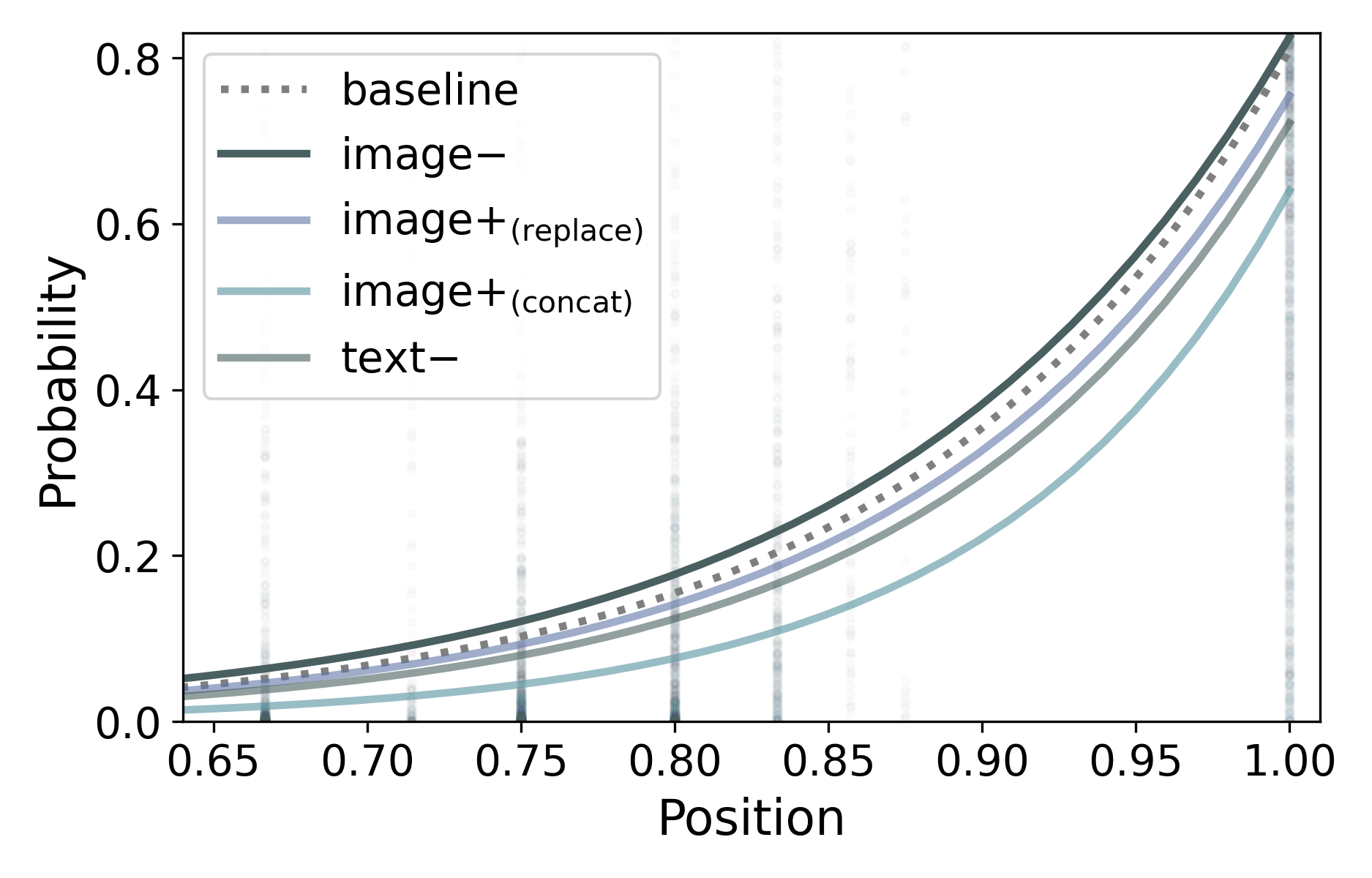}
    \end{center}
    \vspace{-10pt}
\caption{The predictive probability of the EOS token at different target positions within a sequence.}
\label{fig:appendix-context-manipulation}
\end{figure}

%% file: figures/appendix-eos-tendency.tex
\begin{figure}[t]
    \begin{center}
        \includegraphics[width=0.9\linewidth]{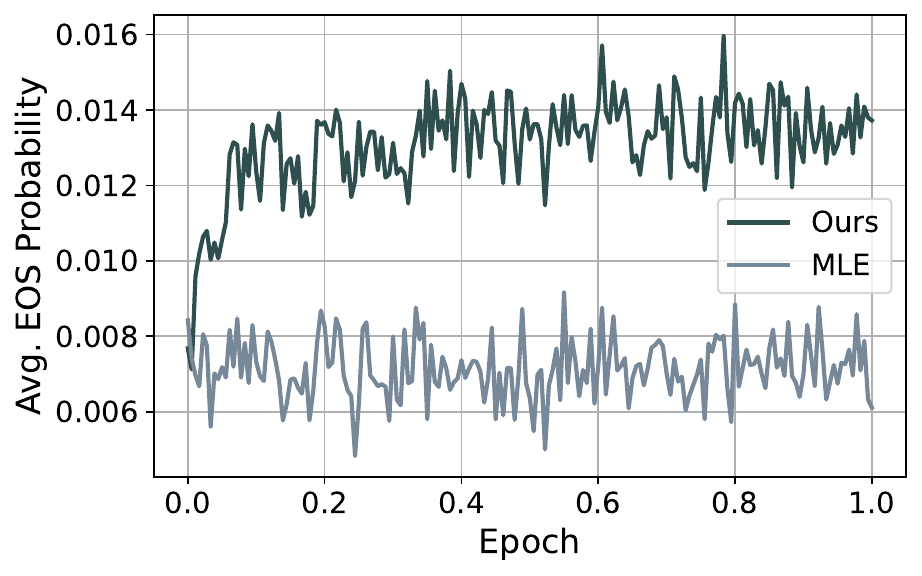}
    \end{center}
    \vspace{-8pt}
\caption{The average probability of the LLaVA-1.5 {\scriptsize{(7b)}} model predicting the EOS token at each position within the minibatch during further training.}
\label{fig:appendix-eos-tendency}
\end{figure}

%% file: tables/appendix-omitted.tex
\begin{table}[t]
\caption{\label{tab:omitted}
Hallucinated and correct objects ``omitted'' from the original model outputs by our methods.
}
\vspace{-4pt}
\centering
\small
\begin{tabular}{@{}l|ccc@{}}
\toprule
Method & \#Halluc. & \#Correct & Halluc. Rate↑ \\ \midrule
\textbf{Ours \scriptsize{(w/ Inst.)}} & 263 & 104 & 71.7\% \\
\textbf{Ours \scriptsize{(w/ Cap.)}} & 244 & 93 & \textbf{72.4\%} \\ \bottomrule
\end{tabular}
\end{table}

\begin{table}[!ht]
\caption{\label{tab:omitted-2}
Average correct and hallucinated object counts of generated captions. \textbf{Original model}: LLaVA-1.5 \scriptsize{(7b)}.
}
\vspace{-4pt}
\centering
\small
\begin{tabular}{@{}l|cc@{}}
\toprule
Model & \#Correct↑ & \#Hallucinated↓ \\ \midrule
Original model & \textbf{2.45} & 0.90 \\
\textbf{Ours \scriptsize{(w/ Cap.)}} & 2.40 & 0.63 \\
\textbf{Ours \scriptsize{(w/ Inst.)}} & 2.36 & \textbf{0.55} \\ \bottomrule
\end{tabular}
\end{table}

%% file: figures/appendix-score-distribution.tex
\begin{figure}[t]
    \begin{center}
        \includegraphics[width=0.9\linewidth]{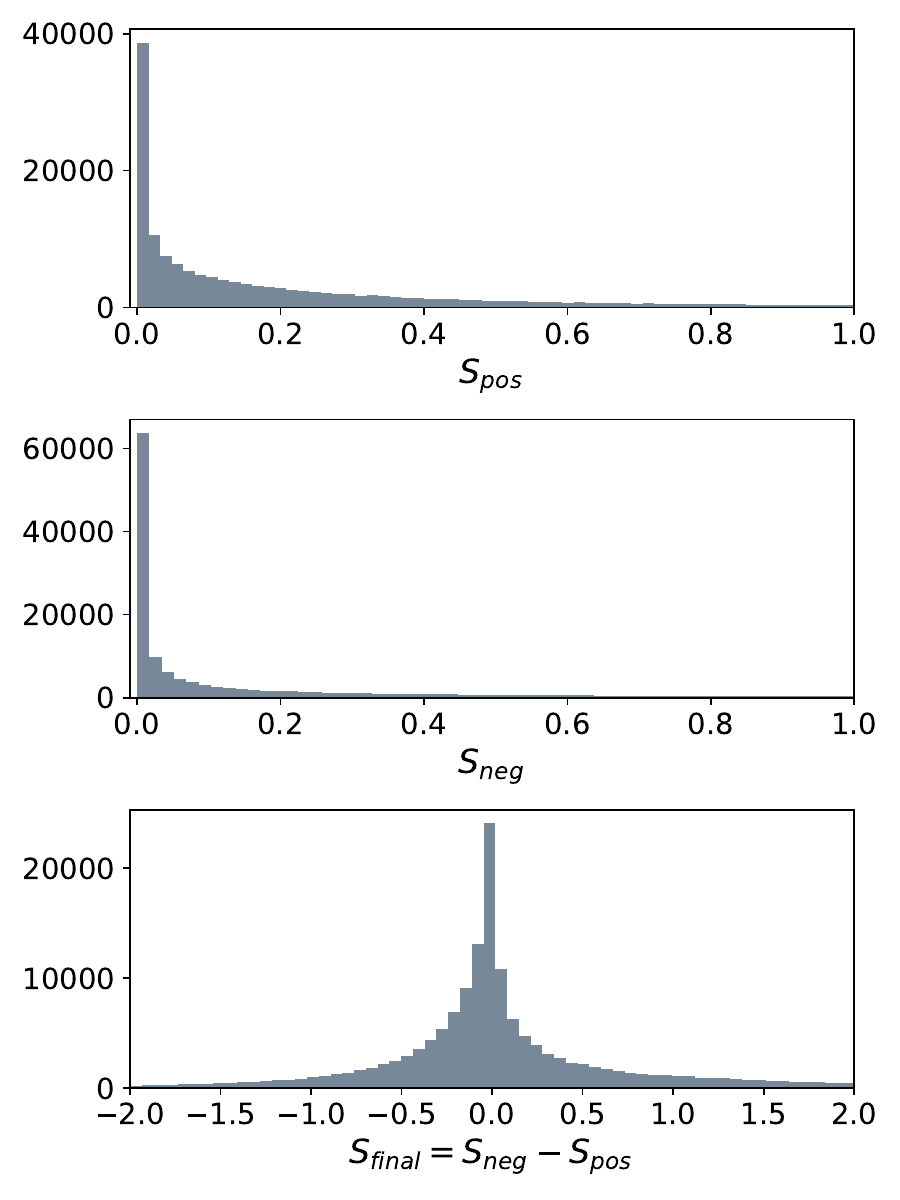}
    \end{center}
    \vspace{-8pt}
\caption{The score distributions of \(S_{pos}\), \(S_{neg}\), and \(S_{final}\) in the LLaVA-Instruction-150K dataset.}
\label{fig:appendix-score-distribution}
\vspace{-8pt}
\end{figure}

%% file: tables/appendix-mme-pope.tex
\begin{table*}[t]
\caption{\label{tab:appendix-mme-pope}
MME and POPE evaluation results of baselines and models trained with our proposed two methods. For LLaVA-1.5 {\scriptsize{(7b)}}, we compare the original model (\textbf{Baseline}), the model trained with \textbf{MLE}, and the one with Selective EOS Supervision (\textbf{Ours}). For LLaVA {\scriptsize{(7b)}}, \textbf{Baseline} and \textbf{Ours} refer to the models trained with the original data and the data filtered by our Scoring EOS Supervision, respectively.
}
\vspace{-4pt}
\centering
\small
\begin{tabular}{@{}l|c|cc|cccc@{}}
\toprule
\multirow{2}{*}{Model} & \multirow{2}{*}{Method} & \multicolumn{2}{c|}{MME} & \multicolumn{4}{c}{POPE} \\ \cmidrule(l){3-8} 
 &  & Perception & Cognition & F1 & Accuracy & Precision & Recall \\ \midrule
\multirow{3}{*}{LLaVA-1.5 {\scriptsize{(7b)}}} & Baseline & 1,516.1 & 348.2 & 85.9 & 86.9 & 94.0 & 79.1 \\
 & MLE & 1,470.9 & 372.5 & 86.1 & 87.0 & 93.6 & 79.7 \\
 & Ours & 1,490.4 & 367.9 & 86.0 & 86.8 & 93.5 & 79.5 \\ \midrule
\multirow{2}{*}{LLaVA {\scriptsize{(7b)}}} & Baseline & 883.1 & 263.6 & 73.3 & 63.8 & 58.8 & 97.5 \\
 & Ours & 910.9 & 260.0 & 71.2 & 59.7 & 56.0 & 98.1 \\ \bottomrule
\end{tabular}
\end{table*}

%% file: figures/appendix-case.tex
\begin{figure*}[ht]
    \begin{center}
        \includegraphics[width=1\linewidth]{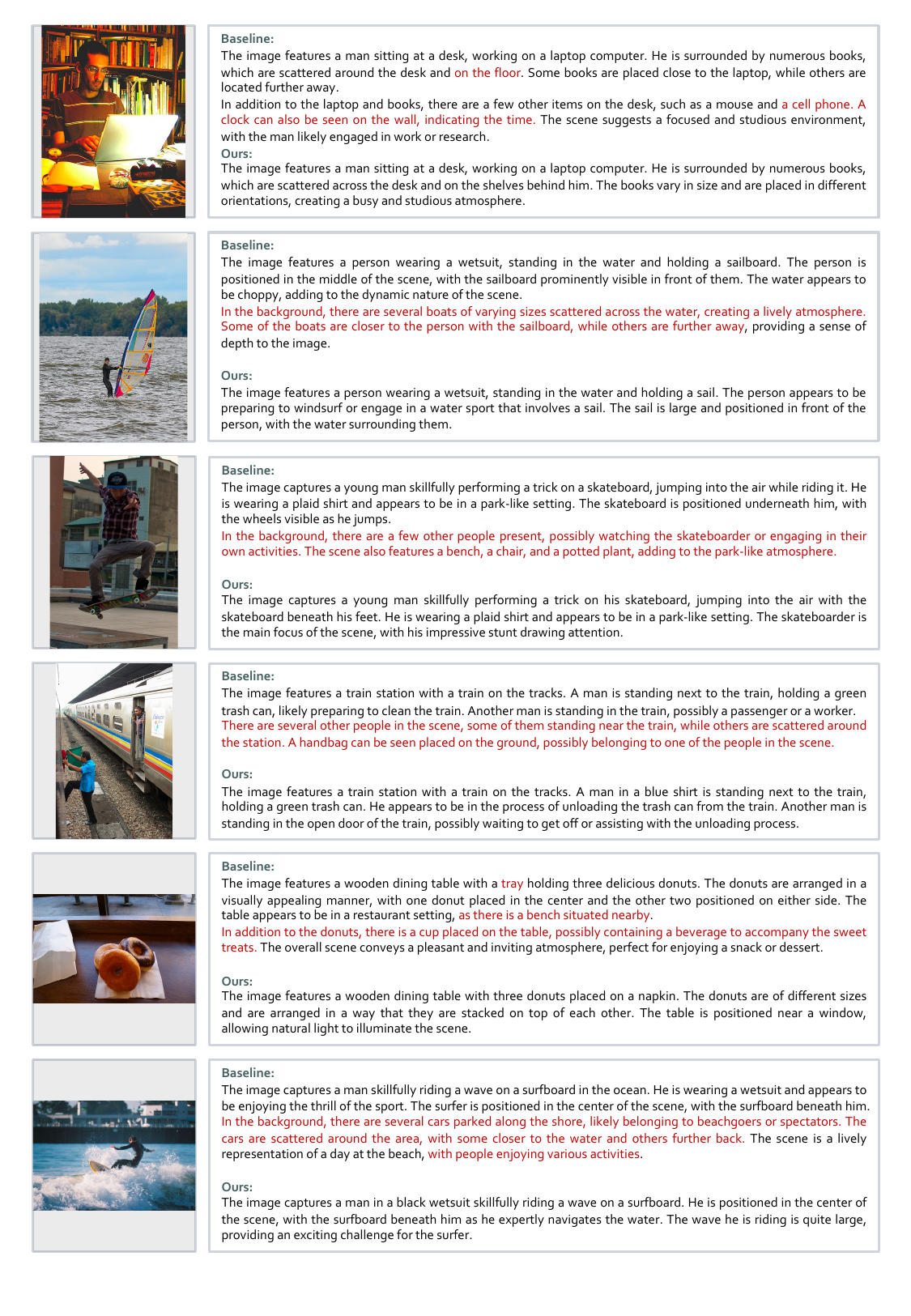}
    \end{center}
    \vspace{-8pt}
\caption{Qualitative results of the LLaVA-1.5 {\scriptsize{(7b)}} model (\textbf{Baseline}) and its counterpart further trained on LLaVA-Instruction-150K with Selective EOS Supervision (\textbf{Ours}).}
\label{fig:appendix-case-1}
\end{figure*}

\begin{figure*}[ht]
    \begin{center}
        \includegraphics[width=1\linewidth]{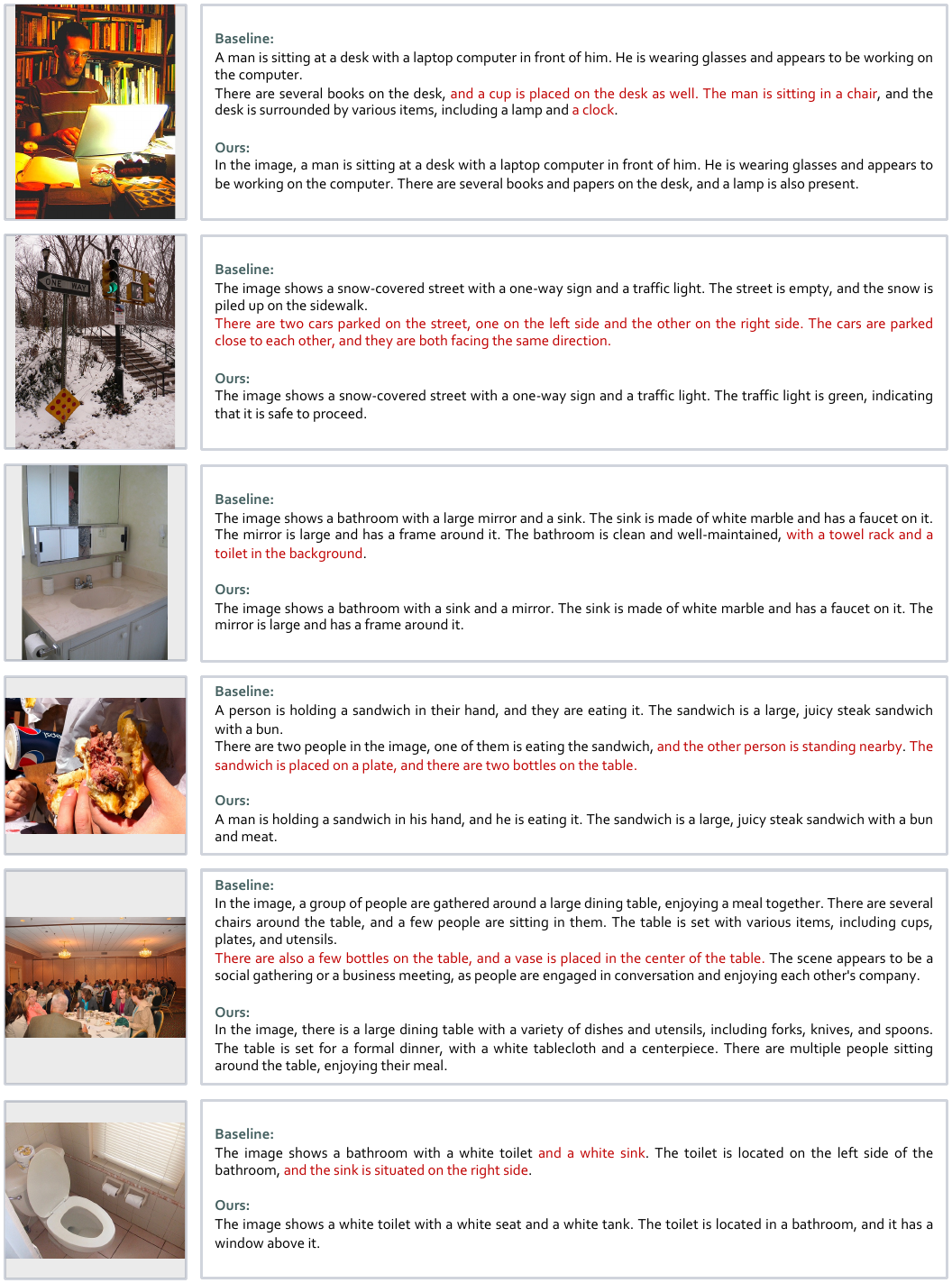}
    \end{center}
    \vspace{-8pt}
\caption{Qualitative results of the LLaVA {\scriptsize{(7b)}} model trained with original LLaVA-Instruction-150K data (\textbf{Baseline}) and with the data filtered by Scoring EOS Supervision (\textbf{Ours}).}
\label{fig:appendix-case-2}
\end{figure*}